# Feature-Refined Unsupervised Model for Loanword Detection

Promise Dodzi Kpoglu
*LLACAN, CNRS*

**Abstract**

We propose an unsupervised method for detecting loanwords i.e., words borrowed from one language into another. While prior work has primarily relied on language-external information to identify loanwords, such approaches can introduce circularity and constraints into the historical linguistics workflow. In contrast, our model relies solely on language-internal information to process both native and borrowed words in monolingual and multilingual wordlists. By extracting pertinent linguistic features, scoring them, and mapping them probabilistically, we iteratively refine initial results by identifying and generalizing from emerging patterns until convergence. This hybrid approach leverages both linguistic and statistical cues to guide the discovery process. We evaluate our method on the task of isolating loanwords in datasets from six standard Indo-European languages: English, German, French, Italian, Spanish, and Portuguese. Experimental results demonstrate that our model outperforms baseline methods, with strong performance gains observed when scaling to cross-linguistic data.

## 1. Introduction

Languages change constantly guided not only by internal mechanisms but also through interactions with other languages. One of the most visible effects of such contacts is when words are transferred from one language into another. Understanding this process in a language can reveal insights into the historical trajectories of various components of the language. For instance, speakers of Spanish in the Caribbean, following Columbus's expeditions, acquired and passed words such as *tobacco*, *canoe*, *hammock*, *potato* and *barbecue* from Taino, an Arawakan language, into Spanish - these words subsequently entered English (Grant, 2015), and understanding this process can give an insight into the relationships that existed between indigenes and the newly arrived populations.

Words that are borrowed from one language into another are called loanwords, and they can possess recognizable features and traceable paths of diffusion. These features and diffusion paths can be identified and tracked by linguists so as to reconstruct the nature and timing of language contact. In the field of historical linguistics, such a task is critical as it allows linguists to distinguish between cases of contact-induced resemblance between words of different languages and genetic inheritance (Grant, 2000). The consequence of this is that more refined phylogenetic models of language evolution can be proposed (List, 2019b). Thus, in computationally elaborating phylogenetic models of language evolution, it is fundamental to isolate loanwords so as not to unnecessarily bias the model (List, 2024) .

Despite their centrality, especially in work related to historical linguistics, the automatic detection of loanwords remains a difficult task in computational historical linguistics (List, 2019a). While recent years have seen modest advances in other related historical linguistics tasks, autonomous unsupervised methods for detecting loanwords seem to have lagged behind.

This paper presents a hybrid, autonomous unsupervised feature-rich model that combines probabilistic analysis with structural indicators, in the detection of loanwords in both monolingual and multilingual contexts, for integration into a historical linguistics workflow. Using a dataset of wordlists from English, French, German, Spanish, Italian and Portuguese, we benchmark the model's performance against a similar method. Experimental results show that our model outperform the baseline model in both precision and recall, with scalable versions also revealing promising results.

Thus, the contributions of this study are threefold:

- Innovation: we present an autonomous unsupervised loanword detection algorithm that integrates feature-modelling and statistical compilation into a probability-like composite scoring mechanism anchored within an iterative refinement space.
- Methodological integration: The system bridges the gap between theoretical work and practical applications, combining linguistic knowledge with statistical compilation for a data-driven discovery.
- Scalability and extensibility: The proposed architecture paves the way for extensions that could incorporate new components for optimal performance.

Thus, this paper contributes a novel, linguistically-informed computational approach to loanword detection. It offers a pathway towards more autonomous, accurate and scalable models of loanword detection, especially within historical linguistics workflows. We release our code at https://github.com/PromiseDodzi/loanword_detection.

## 1.1. Background

### 1.1.1. What are loanwords?

The term *loanword* is frequently used interchangeably with *lexical borrowing*, though the two are not strictly synonymous. Following Haspelmath (2009), *loanwords* can be defined more narrowly as lexical items that are directly transferred from one language into another. This definition excludes broader borrowing phenomena such as semantic calques, syntactic structures, or pragmatic conventions, which go beyond the direct importation of lexical forms. In this sense, loanwords are distinguished from *native words*, which are typically regarded as having evolved internally within a language without external lexical influence (Lehmann, 1962). For example, in English, the word *career* can be analyzed as a loanword, since it is a borrowing from French into English. In contrast, the word *full* is a native English word, which can be traced through Old English (*full*), Proto-Germanic (*fullaz*), and ultimately to Proto-Indo-European (*ple-*), representing a chain of internal linguistic development.

Once incorporated into the recipient language, loanwords undergo different degrees of linguistic integration. This integration can occur along multiple dimensions, including phonological, morphological, and syntactic adaptation. The degree of integration may range from a relatively faithful preservation of the original form to extensive modifications that align the loanword with the structural norms of the recipient language. Hafez (1996) for instance, provides an in-depth analysis of the integration processes of English loanwords into Egyptian Arabic, demonstrating various patterns of phonological and morpho-syntactic integration.

It can be suspected that the extent of integration of loanwords into the recipient language correlates with the chronological depth of borrowing i.e., older loanwords are presumed to be more fully integrated into the recipient language than more recent ones. However, empirical studies challenge such an assumption. Works such as Poplack & Dion (2012) on the morpho-syntactic integration of English loanwords in Quebec French, and Paradis & LaCharité (2008) on phonological adaptation, reveal that the degree of integration does not consistently align with the age of the loanword in the recipient language. These findings suggest that integration is influenced by multiple linguistic and extralinguistic factors, rather than simply by the time elapsed since borrowing.

### 1.1.2. Features of loanwords

Loanwords manifest a range of linguistic features that differentiate them from native words. Grant (2000) identifies two major processes involved when loanwords enter the recipient language: *fabric transfer* and *pattern transfer*. Fabric transfer includes phonetic and phonological characteristics that accompany the borrowed word into the recipient language. Pattern transfer, on the other hand, refers to the carry-over of syntactic, semantic, or pragmatic properties. While fabric transfer is typically an unmarked feature of loanwords, pattern transfer is variable and not always present. In other words, where features characterize a loanword, there are always enough phonological traces "foreignness", while the availability of traces at other linguistic levels i.e., morphological, syntactic, or semantic, may be inconsistent.

A factor to be noted is that these key linguistic features that characterize loanwords (fabrics and patterns) are independent of the time depth of their integration. Crucially however, these features often serve as diagnostic indicators of their foreign origin, with phonological adaptations being the most salient. Among these features are:

- PHONETIC/PHONOLOGICAL FEATURES: As noted above, loanwords often undergo phonological adaptation upon entering the recipient language. Kang (2011) identifies five major patterns of adaptation: too-many-solutions patterns, divergent repair patterns, unnecessary repair patterns, retreat to the unmarked patterns, and differential importation patterns, and these pattern types are central to theoretical debates in linguistics. More importantly, because segmental adaptations are nearly universal across borrowing scenarios, any analytical framework, whether theoretical or algorithmic, must adopt a holistic approach that accounts for the full range of emergent patterns.

- PHONOTACTIC FEATURES: Loanwords can violate the native phonotactic constraints of the recipient language, often introducing otherwise illicit phoneme combinations or syllable structures. For example, Crawford (2009) notes instances in Japanese where loanwords introduce onset clusters or vowel sequences not permitted in native words. These violations can persist across generations, demonstrating that loan phonotactics can override native markedness constraints irrespective of time depth of entry into the recipient language, thereby further highlighting their diagnostic utility.

- MORPHOLOGICAL FEATURES: The morphological integration of loanwords may follow regular inflectional and derivational paradigms in the recipient language, but not without some idiosyncrasy. Ralli & Rouvalis (2022), in their study of Kaliardá (a Greek-based antilanguage), observe that while many loanwords conform to expected morphological patterns, others resist such regularization and display peculiar behavior. This variability is illustrative of the complex nature of morphological adaption that can be characteristic of loanwords.

- SYNTACTIC FEATURES: Loanwords tend to enter specific grammatical categories more readily than others. It often is the fact that nouns are significantly more borrowable than verbs, adjectives, or function words. Tadmor et al., (2010), in a typological survey, report that 31.2% of analyzed nouns across languages are of foreign origin, compared to only 14.0% of verbs. Adjectives and adverbs also exhibit low borrowability (~15.2%), while function words such as pronouns, prepositions, and conjunctions are the least likely to be borrowed. This distribution suggests a general tendency for loanwords to enter the open-class lexical categories rather than closed-class grammatical categories.

- SEMANTIC FEATURES: The semantic domains of borrowings also reveal consistent patterns. Since Swadesh (1955), it has been understood that certain lexical items form a "core" vocabulary that is resistant to borrowing. Words such as *mother*, *hand*, *run*, *sleep*, *sun*, and *I* are generally considered part of this core and thus less likely to be borrowed. Although the precise composition of core vocabulary is still debated, it is widely agreed that borrowings tend to cluster around non-core concepts, particularly those related to technology, culture, and innovation (Thomason & Kaufman, 2001). From a cross-linguistic perspective, this semantic tendency and its relationship to borrowing provides further clues for identifying lexical borrowings.

## 2. Related work

Approaches to the automatic detection of loanwords generally fall into two broad categories: supervised and unsupervised models. Supervised models involve both traditional machine learning algorithms and more recent neural network-based architectures. For example, Miller et al., (2020) applied classic classification techniques to detect loanwords, while Alvarez-Mellado et al., (2025) developed neural models with sophisticated architectures aimed at improving detection accuracy. However, within the context of historical linguistics, supervised models may not always be the most appropriate tools. As List (2024) argues, tasks in historical linguistics often extend beyond the capabilities of supervised inference methods, requiring models that can, without supervision, engage with the complexities of diachronic linguistic phenomena.

In contrast, unsupervised models for loanword detection leverage various language features to identify likely candidates. According to List (2019a), such techniques generally fall into three broad categories, each exploiting a different source of information: phylogenetic conflict, form similarity, and semantic divergence. Among these, phylogenetic conflict is often the most immediately accessible. This approach is grounded in the assumption that if two languages belonging to distinct families exhibit high similarity in certain word forms, it is more plausible that these word forms are loanwords rather than words of common descent i.e., cognates (List, 2019b). This methodology has proven effective in cases where the languages in question are genealogically unrelated.

However, it is important to note that borrowing also occurs between genetically related languages, which significantly complicates the task of loanword detection. For example, although English and French belong to the Germanic and Romance subfamilies, respectively, the English word *mountain* is borrowed from the French *montagne*. In such cases, the formal similarities between the words, especially as the languages belong to different sub-families, makes the loanword detection relatively easy. On the other hand, the French word *bravo* is borrowed from Italian, although both languages belong to the Romance family. In such an instance, the phylogenetic information is uninformative to loanword detection, and such loanwords can be missed. More importantly, even when phylogenetic information is pertinent, there is a risk classifying words of common descent across languages of different sub-families as loanwords (*name* and *nom* in English and French respectively).

Given these inherent challenges, the most robust unsupervised approaches often combine two or more factors subtly. Hantgan & List (2022), for instance, analyze the discrepancy in cognate identification between different automated methods, such as LexStat (List, 2012a) and Sound-Class-Based Alignment (SCA; List, 2012b), as a clue to identifying potential loanwords - it is worthy to note that despite the task being loanword detection, their task also involved layered contacts (contact between languages can manifest in various ways in the lexicon, thus theoretically going beyond mere loanword detection). Similarly, Mennecier et al., (2016) examine similarity thresholds across languages from different families, attempting to isolate patterns indicative of loan events.

Despite their relative success, these approaches are fundamentally biased toward *language-external* data. That is, they rely heavily on comparisons with other languages or families before making determinations about loanword status. In phylogenetic methods, genealogies are consulted; in similarity-based approaches, linguistic data from other languages are analyzed; and in semantic models, the core/non-core vocabulary distinction is often leveraged to identify non-native lexical items. Therefore, these methods can be described as non-autonomous. Indeed, these non-autonomous methodologies have yielded impressive results. For example, List & Forkel (2022) reported a precision of 0.94, recall of 0.81, and F1-score of 0.87 for loanword detection in Southeast Asian languages.

Nevertheless, non-autonomous methods are subject to several notable constraints. First, their efficacy may be diminished when applied to datasets comprising languages within the same family, as the high degree of structural and lexical similarity can obscure signals of 'loanwordedness'. Second, these methods can presuppose the availability of accurate phylogenetic information, despite the fact that phylogenetic relationships between groups of languages may be conceptualized differently by different linguists, thereby introducing potential inconsistencies into model outputs. For example, the internal phylogeny of the Dogon language cluster of Mali remains unresolved (Heath, 2022; Moran & Prokić, 2013). Furthermore, because one of the principal motivations for automatic borrowing detection in historical linguistics is to support fine-grained phylogenetic reconstruction (Rama & List, 2019), such reliance on pre-existing phylogenetic frameworks risks introducing a form of methodological circularity. Consequently, these non-autonomous approaches are simultaneously constrained by intra-family similarity effects and by potential methodological circularity.

In light of these challenges, our work proposes an alternative approach that emphasizes *language-internal* features. Instead of relying on cross-linguistic comparisons or external phylogenetic models, we focus on identifying internal featural conflicts that may signal the presence of loanwords. It should be noted that this strategy is not without its own difficulties. As noted earlier, loanwords often undergo varying degrees of integration, ranging from highly foreign forms to those that are nearly indistinguishable from native words, making autonomous detection particularly complex. Nonetheless, prior work has demonstrated that language-internal features can be harnessed effectively. For example, Prakhya & P (2020) employed the syllabic properties of Malayalam and Telugu to detect loanwords and reported promising results.

Our proposed model builds upon these foundations by extending beyond syllabic features to incorporate a broader set of linguistic cues. Specifically, it integrates phonotactic, morphological and syntactic information in a combinatorial manner. By leveraging this multi-feature, language-internal framework, our model aims to enhance the precision and reliability of automatic loanword detection, while avoiding the pitfalls associated with external assumptions. Experiments show that our model outperforms the existing baseline, with a scalable version which allows for the incorporation of unconstrained cross-linguistic information (external-information) resulting in even more promising results. Thus, the current model is not only innovative, but also complements existing autonomous and non-autonomous models.

## 3. Task description

Loanword detection involves isolating words of foreign origin from the native words of a language. Therefore, given a set of words of a language i.e., a vocabulary, the task is to predict which forms are loanwords without any prior training data. Consequently, unlike in models that access language-external information, our model takes as input a monolingual wordlist. The assumption is that, features that are characteristic of native words are largely different from those that are characteristic of loanwords. We nevertheless admit that native word features and loanword features, at a singular linguistic level can be non- distinctive. The goal then is to call on a comprehensive set of cues across various linguistic levels, and model them probabilistically for the determination of words likely to be loanwords.

Beyond the monolingual wordlist, language-internal information captured this way can be combined with language-external information, in a scalable variant of the model, in order to further improve performance. Thus, while our basic model operates strictly on the monolingual wordlist, the scalable version allows for the integration of non-constraining language-external information i.e., cross-linguistic information that is not constrained by languages belonging to the same family.

The performance of the basic model i.e., the model that operates on a strictly monolingual wordlist, is evaluated against a reimplementation of the model introduced by Prakhya & P, (2020), allowing for a direct comparison in a similar unsupervised setting.

Formally, let $\Sigma$ represent the total phonemes of a language, and let $V \subseteq \Sigma^*$ denote the vocabulary, i.e., the set of words on which loanword detection is to be performed. The objective is to define a function: $f : V \rightarrow \{0,1\}$ such that for any word $w \in V$, $f(w) = 1$ if $w$ is predicted to be a loanword and $f(w) = 0$ if $w$ is predicted to be a native word.

The task is unsupervised, so no labeled data is available for training. Instead, $f$ is inferred from features observable within $V$ itself, guided by the hypothesis that native and loanwords exhibit statistically distinguishable patterns across multiple linguistic levels. To make this possible, a scoring function $s: V \rightarrow R$ is used, such that higher values of $s(w)$ indicate a higher likelihood of $w$ being a loanword. The detection function $f$ can then be defined as: $f(w) = 1$ if $s(w) \geq \tau$; 0 otherwise, where $\tau$ is a threshold to be determined empirically.

In the basic model, $s(w)$ is derived entirely from language-internal features. In the scalable variant, $s(w)$ is further augmented with cross-linguistic signals which takes the form of cross-linguistic similarity computed by comparing $w$ against wordlists from other languages. This then defines a multi-layered, hybrid and scaled architecture that adapts from strictly monolingual inference.

## 3.1. Dataset

The dataset used for model evaluation consists of a manually curated lexicon comprising 836 unique lexical concepts distributed across six standard Indo-European languages: English, German, French, Spanish, Italian and Portuguese. For each lexical item, the dataset includes its orthographic form, International Phonetic Alphabet (IPA) transcription, language, part-of-speech, and a binary classification label indicating loanword status. The data are organized in tabular format with one row per lexical entry, resulting in a total of 5,092 data points across all languages.

| LANGUAGE | NATIVE WORDS | LOANWORDS |
|---|---|---|
| English | 423 | 414 |
| German | 502 | 382 |
| French | 549 | 292 |
| Spanish | 532 | 311 |
| Italian | 590 | 253 |
| Portuguese | 512 | 332 |

**Table 1.** *Distribution of lexical entries by language and loanword status.*

All experiments were conducted exclusively on IPA transcriptions, rather than on orthographic representations, to ensure cross-linguistic phonological comparability. Prior to processing, IPA strings were normalized by removing length markers (ː) and primary stress markers (ˈ) to eliminate features irrelevant to the task. Moreover, the annotation of loanword status required both documented etymological evidence and principled linguistic judgment. For example, the French word *téléphone* (of Greek origin) was unambiguously classified as a loanword, whereas more complex cases such as *télévision* (a hybrid form with Greek and Latin components) required adjudication. In such cases, consistent labeling criteria were applied, and *télévision* was classified as a loanword. These annotation decisions reflect an effort to balance etymological accuracy with operational consistency.

## 4. Basic model

We propose a basic model for the identification of loanwords by integrating feature extraction and modelling within an iterative refinement framework. The model analyzes IPA transcriptions using cues from phonotactics, morphology and syntax and then scores them before iterating over the results for refinement. Core features include scoring for rare n-grams, compiling transition probabilities, and weighting according to part-of-speech (POS). The model then applies a multi-stage process that separates native words from loanwords using normalized and custom-weighted features, with length normalization and sigmoid-based probability mapping. Furthermore, a pattern-tracking database aids in identifying native and loanword morphemes. This is all then fed into an iteration loop which continues until convergence, with post-processing rules enhancing confident classifications. Following this, the model outputs both probabilities and binary labels per word, with a tuned decision threshold that balances precision and recall.

As the task is unsupervised, the model relies on linguistic features and their statistical correlates to estimate a function. This is achieved by defining a scoring function that is then mapped unto probabilities. Prior to classification, the results are subjected to some iteration until convergence. Thus, three main components can be isolated for the model: a feature extraction module, a scoring engine and a refiner. Figure 1 presents an overview of the architecture of the basic model.

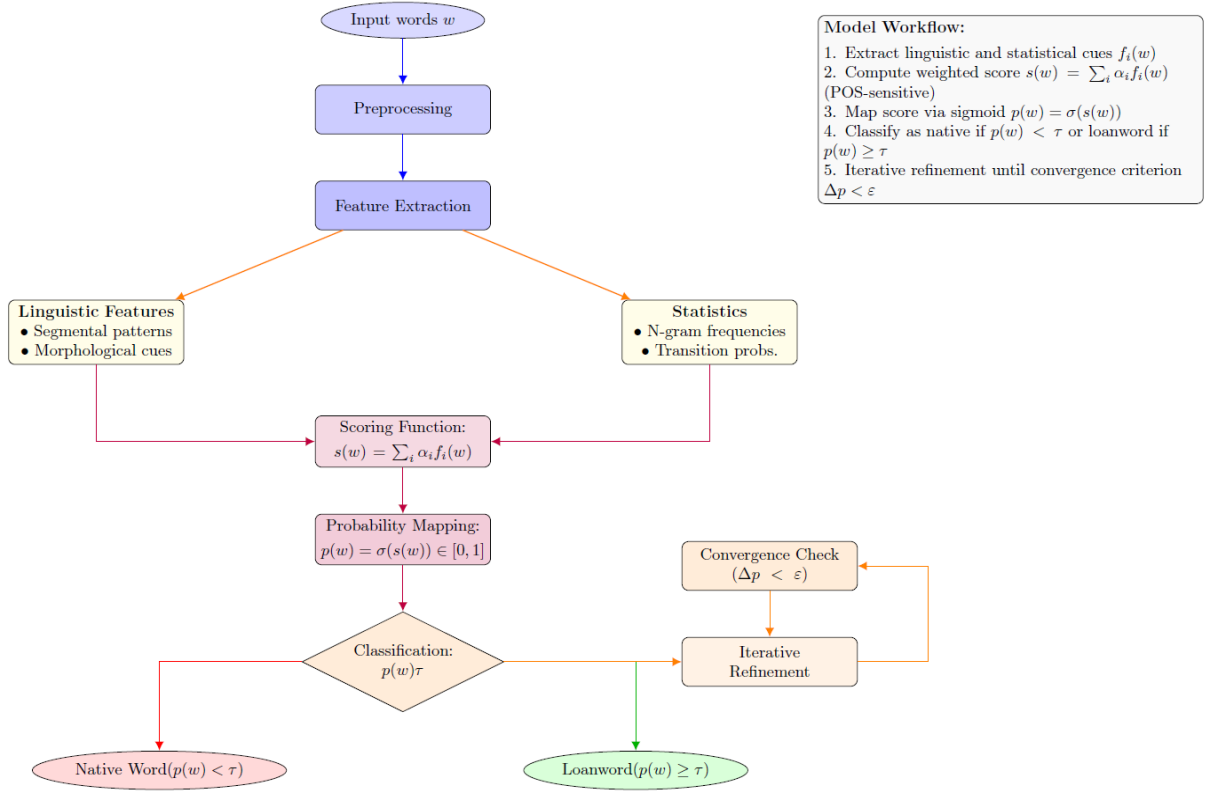

**Figure 1**: *Overview of the basic loanword detection model, which adopts a hybrid approach. The model operates on words of vocabulary and proceeds through a feature extraction phase. Results obtained at this stage are scored and mapped probabilistically to classify words as loans. This is fed into an iteration loop for refinement until convergence occurs.*

### 4.1. Feature extraction

The ability to capture internal contrasts between candidate native words and loanwords depends directly on the discriminative power of the feature extraction module. The linguistic features that are the basis of the model span phonetic, phonotactic, morphological and syntactic properties of the words. These are then built into statistical correlates. They include n-gram based features and transition probabilities.

### 4.1.1. N-gram-based features

Let $n$ n-gram statistics capture local segmental patterns. For each word w ∈ V, its character $n$ n-grams are computed for n ∈ [2,10]. Then, let count($g$) be the frequency of $n$ n-gram $g$ across $V$, and let G(w) be the multiset of $n$ n-grams in $w$. We define a rare n-gram score and an n-gram entropy respectively as:

$$\text{rare\_ngram\_score}(w) = \frac{1}{|G(w)|} \sum_{g \in G(w)} \left[ \mathbb{1}_{p(g) < \epsilon_1} \left( c_1 \cdot (\epsilon_1 - p(g)) \right) + \mathbb{1}_{\epsilon_1 \leq p(g) < \epsilon_2} \left( c_2 \cdot (\epsilon_2 - p(g)) \right) \right]$$

$$H_{\text{ngram}}(w) = - \sum_{g \in G(w)} p(g) \log_2 p(g)$$

The rare n-gram score measures how unusual a word is, based on the rarity of its sub word components, while low entropy from the n-gram entropy score suggests skewed information which are often associated with loanwords.

### 4.1.2. Transition probabilities

Transition between segments can be idiosyncratic for loanwords. Consequently, we extract scores based on this from various perspectives. First, let $P(c_i \to c_j)$ be the empirical bigram transition probability in $V$. We define a rare transition score: $\text{rare\_transition\_score}(w) = \frac{1}{T(w)} \sum_i [\mathbb{1}_{p<0.01} \cdot 100(0.01 - p) + \mathbb{1}_{0.01 \leq p < 0.05} \cdot 20(0.05 - p)]$ where $T(w)$ is the number of character transitions in $w$; a transition entropy: $H_{\text{trans}}(w) = -\sum_i p_i \log_2 p_i$ and an average transition probability: $\text{avg\_trans\_prob}(w) = \frac{1}{T(w)} \sum_i p(c_i \to c_{i+1})$.

Once all the features are extracted, a standardized length normalization component is added. $z_{\text{len}}(w) = \frac{|w| - \mu_L}{\sigma_L}$ where $\mu_L$, $\sigma_L$ are the mean and standard deviation of word lengths in $V$. The idea is that words significantly longer or shorter than the average are potentially non-native. The final output of the feature extractor is a vector $x(w) \in R^d$ for each word $w \in V$, where $d = $ total number of extracted features. Since each feature captures a distinct signal of "borrowedness" these feature vectors form the foundation for scoring and mapping in the scoring engine.

### 4.2. The scoring engine

The scoring engine takes as input the extracted feature vectors from vocabulary and computes a composite score for each word. These scores are then normalized and transformed into probability estimates of being a loanword. Thus, let each word $w \in V$ be associated with a feature vector $x(w) = [x_1(w), x_2(w), ..., x_d(w)] \in R^d$ where each $x_i(w)$ corresponds to one of the $d$ normalized feature values extracted from the feature extractor. The goal of the scoring engine is to convert this high-dimensional representation into a scalar score that reflects the likelihood that $w$ is a loanword.

To accomplish this, we define a composite scoring function $s: V \to R$ that aggregates feature contributions via a weighted linear combination: $s(w) = \sum_{i=1}^{d} \alpha_i \cdot \hat{x}_i(w)$, where $\hat{x}_i(w) \in [0,1]$ is the normalized value of the $i^{th}$ feature, and $\alpha_i \in R$ is the weight assigned to the $i^{th}$ feature, reflecting its relative importance in identifying loanwords. Prior to being fed into the scoring function, each raw feature vector $x_i(w)$ is scaled using min-max normalization over the vocabulary. This ensures that each feature contributes on the same scale, making the weighted sum meaningful.

The final score is adjusted via a length-based modifier implemented using the definition: $\lambda_{\text{len}}(w) = 1 + \left(0.5 \cdot \left(\frac{1}{1+e^{-3(|z_{\text{len}}(w)|-1.5)}} 0.5\right)\right)$ to account for words that are either too long or too short. This nonlinear scaling increases the weight of words with abnormally long or short lengths, amplifying scores for length outliers. Also, final scores are adjusted for syntactic category, which reflects the tendency of certain syntactic categories to be borrowed as against others. Let $\text{POS}(w)$ be the syntactic category of word $w$, e.g., noun, verb, adjective. The syntactic adjustment is implemented as: $\lambda_{\text{pos}}(w) = \beta_{\text{POS}(w)}$ where $\beta_{\text{POS}(w)} \in (0,1]$ is a predefined weight associated with each syntactic category. Nouns, which are more likely to be borrowed, are typically assigned higher weights (e.g., $\beta_{\text{noun}} = 1.0$). The final adjusted score for word $w$ is thus computed as $s'(w) = s(w) \cdot \lambda_{\text{len}}(w) \cdot \lambda_{\text{pos}}(w)$.

To translate the scores $s'(w)$ into loanword probability, we apply a scaled sigmoid transformation: $P(w) = \sigma_\gamma(s'(w)) = \frac{1}{1+e^{-\gamma(s'(w)-\theta)}}$ where $\gamma > 0$ controls the steepness of the sigmoid and $\theta$ is the center of the function (defaulting to 0.5). However, to further refine borderline cases, the scoring engine applies heuristic post-processing. If certain features exceed defined anomaly thresholds, the final probability $P(w)$ is boosted. The final probability is then updated as: $P'(w) = \min\left(P(w) \cdot \prod_{f \in F_{\text{anomaly}}} (1 + \eta_f \cdot \delta_f(w)), 1.0\right)$ where each $\eta_f$ is a feature specific boost and $F_{\text{anomaly}}$ is the set of all high-signal features. This dynamic enhancement step sharpens the decision boundary for words exhibiting multiple atypical characteristics. This final probability $P'(w) \in [0,1]$ is used to determine the binary label via a fixed threshold $\tau$.

### 4.3. Iterative refinement and convergence

In the absence of supervision, the model must rely entirely on internal contrasts within the vocabulary. The scoring engine produces an initial probability estimate $P(w) \in [0,1]$ for each word $w \in V$, using statistics drawn from linguistic features derived from the vocabulary as a whole. However, in early passes, these statistics may be biased by various factors. To address this, the algorithm performs iterative re-estimation, progressively improving the classification function $f: V \to \{0,1\}$ by dynamically updating the reference distributions based on increasingly reliable subsets of native words, in a kind of self-training with soft rebalancing and convergence control. The steps involved are: feature re-extraction, statistical baseline refinement, incorporation of morphological patterns, and the application of smoothing and convergence criteria. The goal is to refine classification predictions based on emerging contrasts between native and loan word patterns

To initialize the iterative process, let $P^{(0)}(w)$ be the initial borrowing probabilities computed using the full vocabulary $V$. A threshold $\tau \in [0,1]$ is applied to obtain an initial set of predicted loanwords: $B^{(0)} = \{w \in V \mid P^{(0)}(w) \geq \tau\}$, $N^{(0)} = V \setminus B^{(0)}$ (initial native word set). At each iteration, the model performs the following steps:

I. Feature re-extraction:
   The feature space is re-centered, as new features are extracted from the current set of candidate native words $N^{(t-1)}$. This includes updated n-gram distributions, transition probabilities, and standard deviation of word lengths.

II. Feature re-estimation:
    New statistical estimations are computed from the current native-biased set and the scoring engine is re-run to produce updated scores and probabilities: $P^{(t)}(w) = \text{sigmoid\_score}(x^{(t)}(w))$.

III. Pattern-based refinement:
     For iterations $t \geq 2$, the model performs a refinement step based on phoneme-level string patterns. Two separate pattern databases are constructed: 1. $\mathcal{P}_N^{(t)}$ (prefixes, suffixes and 3-character segments frequent among $N^{(t-1)}$), 2. $\mathcal{P}_B^{(t)}$ (same, but from $B^{(t-1)}$, the set of words predicted to be loanwords at iteration $t-1$). For each word $w$, its pattern-likeness score with respect to the native database is computed as: $PL_N(w) = $

$\frac{1}{|S(w)|}\sum_{p\in S(w)} \frac{\mathcal{P}_N^{(t)}(p)}{\mathcal{P}_N^{(t)}(p)+\mathcal{P}_B^{(t)}(p)+\varepsilon}$ where $S(w)$ is the set of extracted patterns from $w$, and $\varepsilon > 0$ is a smoothing constant. So then, let $A(w)$ be a boolean predicate that indicates if $w$ is anomalous based on its features. If: $A(w) = 0$ and $PL_N(w)$ is high → downscale $P^{(t)}(w)$; $A(w) = 1$ and $PL_N(w)$ is low → upscale $P^{(t)}(w)$. These rule-based post-adjustments sharpen classification by resolving conflicts between surface structure and feature anomaly.

IV. Averaging for stability:
The current probability is then averaged with previous iterations to stabilize classification: $\overline{P}(w)^{(t)} = \frac{1}{t+1}\left(\sum_{k=0}^{t} P^{(k)}(w)\right)$. This running average serves as the final score used to define loanword sets at each iteration: $B^{(t)} = \{w \in V \mid \overline{P}(w)^{(t)} \geq \tau\}$, $N^{(t)} = V \setminus B^{(t)}$

The loop terminates either after a maximum number of iterations $T_{max}$ is reached, or if the change in the predicted loanword set is sufficiently small. This ensures that the iterative refinement process halts when the classification becomes stable, avoiding overfitting or endless oscillation. The final set of predicted loanwords is: $B^* = B^{(t)}$ for the final $t$. In addition to this, each word is also assigned a final smoothed loanword probability $\overline{P}(w)^{(t)}$ which is used in scaling up the model, when it has to take into account cross-linguistic information.

## 5. Scaled model

The basic model can be scaled by the integration of cross-linguistic information, in a non-constraining manner, since in this case, loanwordedness is not solely computed based on cross-linguistic information. In this case, in addition to borrowing probabilities from the base model – which is strictly language-internally derived, information from words in other languages is factored into the predictions; thus, accepting as input a multilingual wordlist.

In this case, let $x$ be a word in language $\mathcal{L}$, and let $B(x) \in [0,1]$ denote the loanword probability assigned by the base model. To incorporate cross-linguistic information, we first define a comparability score, $C(x) \in [0,1]$, which estimates the phonological similarity of $x$ to its concept-aligned counterparts in other languages. This is then combined with the score from the basic model to obtain a composite score which is guided by a dynamic threshold to predict loanword status.

### 5.1. Comparability and composite scoring

The comparability score $C(x) \in [0,1]$ quantifies the phonological dissimilarity of a word $x$ from its cross-linguistic counterparts within the same concept. So, given a concept $c$, let $W_c = \{x_1, x_2, \ldots, x_m\}$ denote the set of IPA-transcribed words associated with that concept across $m$ different languages. The comparability score $C(x) \in [0,1]$ for a word $Wcx \in W_c$ quantifies its phonological atypicality compared to the rest of the concept set. This score is computed by combining (i) context-sensitive alignment log-probabilities and (ii) phonological feature-based distances.

Starting with context-sensitive alignment log-probabilities, for each pair of distinct words $Wc(x_i, x_j) \in W_c \times W_c$ where $i \neq j$, a pairwise alignment is computed using a Needleman–Wunsch-style algorithm. Let the resulting alignment be: $\mathcal{LA}(x_i, x_j) = \{(a_k, b_k, \phi_k)\}_{k=1}^{L}$ where: $a_k \in x_i \cup \{-\}$ $b_k \in x_j \cup \{-\}$ are aligned IPA symbols (or gaps), $\phi_k \in \Sigma^4$ denotes the context window surrounding the aligned pair (e.g., 2 symbols to the left and right), and $L$ is the alignment length. Then, for each aligned pair $(a_k, b_k, \phi_k)$, we estimate a contextual match likelihood $P(a_k, b_k \mid \phi_k)$, derived from the co-occurrence counts of such aligned pairs and contexts across the entire dataset. These probabilities are log-transformed and summed: $\log P_{\text{align}}(x_i, x_j) = \sum_{k=1}^{L} \log P(a_k, b_k \mid \phi_k)$. This quantity captures the likelihood of the alignment given attested cross-linguistic phonotactic regularities. It is higher for phoneme pairs and contexts frequently observed in the training data, and lower for atypical alignments.

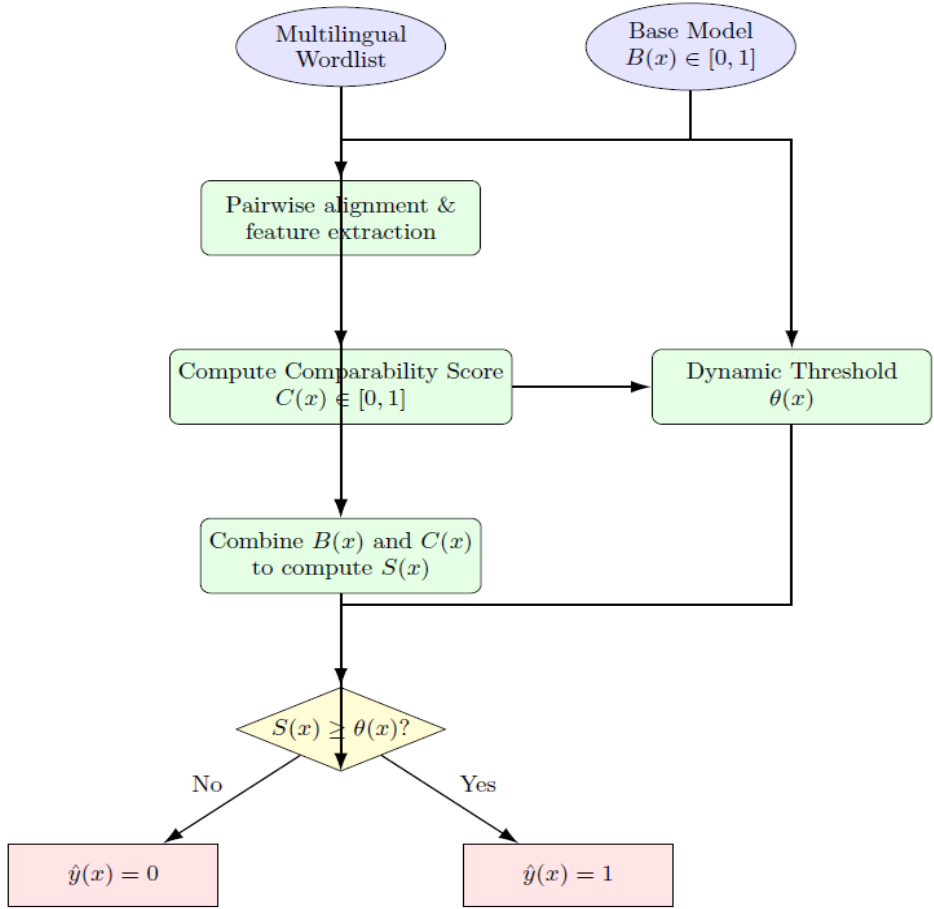

**Figure 2**: *Overview of the scaled loanword detection model, which operates on a composite scoring mechanism. The model accepts words from multilingual wordlist and proceeds through an alignment and feature extraction phase. A comparability score is obtained and combined with borrowing probability. A dynamic threshold allows to classify a word as a loanword or a native word candidate.*

We then continue to get phonological feature-based distances. In this case, each IPA symbol is mapped to a feature vector $df(\cdot) \in R^d$, representing phonological features (e.g., voicing, place, manner, vowel height etc.). The feature edit distance between two aligned characters is: $\text{dist}(a_k, b_k) = \frac{1}{d}\sum_{l=1}^{d} I[f_l(a_k) \neq f_l(b_k)]$. The mean feature distance over the alignment is calculated as: $D_{\text{feat}}(x_i, x_j) = \frac{1}{L}\sum_{k=1}^{L} \text{dist}(a_k, b_k)$. This reflects the articulatory dissimilarity between aligned phonemes.

The total divergence between $x_i$ and $x_j$ is computed as a weighted combination of phonological dissimilarity and contextual improbability: $D(x_i, x_j) = \lambda \cdot D_{\text{feat}}(x_i, x_j) - (1 - \lambda) \cdot$ $\frac{1}{L} \log P_{\text{align}}(x_i, x_j)$, where: $\lambda \in [0,1]$ is a hyperparameter balancing phonological distance and alignment probability, the negative log-likelihood term penalizes improbable alignments, and the division by $L$ ensures length normalization. The comparability score $C(x)$ for word $cx \in W_c$ is thus defined as its average divergence from all other words in the concept set. This yields a scalar in $[0, \infty)$ where higher values indicate greater divergence from the phonological and contextual norms of the concept. Finally, to constrain $C(x)$ to $[0,1]$ and ensure comparability across concepts, we apply min-max normalization within each concept. This score is then integrated with the loanword probability $B(x) \in [0,1]$ obtained from the basic model in a composite scoring function for final prediction.

We the define a composite score $S(x) \in [0,1]$ as a weighted combination of the basic borrowing probability and the inverse of the comparability score: $S(x) = \frac{w_1 B(x) + w_2 (1 - C(x))}{w_1 + w_2}$ where: $w_1, w_2 \in R_{>0}$ are hyperparameters balancing the influence of the base model and the cross-linguistic component. To convert the composite score into a final binary borrowing prediction, we use a dynamic threshold $\theta(x)$ defined as: $\theta(x) = \alpha + \beta \left( (1 - C(x)) - B(x) \right)$ where $\alpha, \beta \in R$ are empirically determined constants controlling the sensitivity to discrepancies between cross-linguistic evidence and the basic model's prediction. Finally, the predicted borrowing label $\hat{y}(x) \in \{0,1\}$ is computed as: $\hat{y}(x) = 1$ if $S(x) \geq \theta(x)$, 0 otherwise.

## 6. Experiments

### 6.1. Baselines

To establish a meaningful performance benchmark for our basic model, we compare it against a reimplementation of the model proposed by Prakhya & P (2020). While our reimplementation maintains the original model's architecture and inference mechanism, we introduce a key modification to accommodate the linguistic scope of our study. The original model was tailored to languages like Malayalam and Telugu, where morphologically meaningful stems may differ from those in the languages we investigate.

To adapt the model, we redefine the notion of a stem: we extract the first two syllables of each word as the stem. This syllabic segmentation is done based on vowel-consonant transitions. For each word, we identify its stem as the concatenation of its first two syllables, and treat the remaining syllables as the "following sequence" used to compute diversity-based nativeness scores. These scores are initialized based on the number of unique syllables that follow a given stem across the vocabulary, with higher diversity indicating nativeness. This adjustment ensures that our baseline remains faithful to the original model's spirit, while better aligning it with the morphological characteristics of our dataset. Thus, our reimplemented model serves as a suitable and fair reference point for evaluating the relative performance of our proposed model.

### 6.2. Ablation studies

Our basic model is built on adequate feature extraction which is scored and modelled in an iterative refiner to predict loanword candidates. For purposes of recall, three types of features are extracted: n-gram-based features, transition probabilities and length-based features. To isolate the contribution of each of these features, we compare the full model $\mathcal{M}$ to variants where individual features are ablated, and one to which new features are added. Specifically, we evaluate $\mathcal{M}-\mathcal{P}1$, a model without n-gram-based features i.e., all n-gram counting and feature extraction is removed (rare n-gram score and n-gram entropy). In the $\mathcal{M}-\mathcal{P}2$ model, transition probabilities are removed, therefore transition entropy, rare transition score and average transition probabilities are not fed into the scoring machine. Also, all transition post-processing rules are eliminated. Finally, the $\mathcal{M-P}3$ model introduces new features which operate at the segmental level into the model: consonant-vowel patterns, character distribution anomalies, consonant clusters and vowel ratios. For the $\mathcal{M-P}3$ model, the objective is to show that increased number of features does not necessarily translate into better loanword detection, and thus the performance of our model is due to feature quality and architectural novelty, especially in the scoring engine and the iterator. In all ablated models, weights of the remaining features are increased to compensate for removing features, while maintaining the same relative importance between them. Each variant is benchmarked on the same reconstruction task, and average classification accuracy is reported using the standard adopted metrics.

### 6.3. Evaluation metrics

To assess the performance of the basic loanword prediction model, a set of standard classification evaluation metrics was employed. The model's predictions were evaluated against a manually annotated gold standard, with each entry in the wordlist labeled as either a loanword (1) or native word (0). From these data, the following classification counts were obtained:

- True Positives (TP): Loanwords correctly identified as loanwords.
- True Negatives (TN): Native words correctly identified as native words.
- False Positives (FP): Native words incorrectly predicted as loanwords.
- False Negatives (FN): Loanwords incorrectly predicted as native words.

These counts formed the basis for calculating the core evaluation metrics:

- Precision: The proportion of predicted loanwords that are correctly identified, calculated as TP / (TP + FP). High precision indicates that the model makes few false positive errors.
- Recall: The proportion of actual loanwords that the model correctly detects, calculated as TP / (TP + FN). A high recall means that the model successfully captures most of the true loanwords, reducing the number of false negatives.
- F1 Score: The harmonic mean of precision and recall, provides a balanced summary of the model's performance.

Given that the loanword detection model is designed to support research in historical linguistics, a high recall is especially desirable. In such a setting, the model serves primarily to suggest candidate loanwords, which are then manually verified by historical linguists. Thus, maximizing recall ensures that fewer genuine loanwords are overlooked, enabling linguists to inspect a broader set of plausible candidates. However, this emphasis on recall does not mean that precision should be disregarded, as overly noisy predictions can reduce the model's practical utility and increase the burden of manual verification. This evaluation approach attempts therefore to balance comprehensiveness with reliability.

### 7. Results and Discussion

Table 2 shows the performance of our basic autonomous borrowing detection model on the data for all the languages, i.e., predictions are done separately on monolingual datasets, and a combined metrics is calculated. This is compared to the baseline UNS model of Prakhya & P (2020) and 2 ablated versions of our basic model in which n-gram-based features (AUT_BOR_P1), and transition probabilities (AUT_BOR_P2) are not taken into account respectively. Finally, we also construct an augmented model in which we extend the features to include various segmental features (AUT_BOR_AUG).

Beyond precision, our basic loanword detection model demonstrates superior performance compared to the baseline and all other tested variants with respect to recall and F1 score. Given that the intended application of the model is within the workflow of historical linguists, the primary objective is to minimize false negatives while maintaining an acceptable, though not minimal, level of false positives. Under this criterion, our model achieves the most favorable trade-off among all evaluated systems.

|  | METRICS | | | CONFUSION MATRIX | | | |
| --- | --- | --- | --- | --- | --- | --- | --- |
|  | PRECISION | RECALL | F1 | TP | FP | TN | FN |
| AUT_BOR | 0.63 | **0.71** | **0.67** | 1400 | 822 | 2286 | 584 |
| AUT_BOR-P1 | 0.59 | 0.17 | 0.26 | 336 | 229 | 2879 | 1648 |
| AUT_BOR-P2 | **0.66** | 0.51 | 0.57 | 1001 | 508 | 2600 | 983 |
| AUT_BOR-AUG | 0.64 | 0.58 | 0.61 | 1151 | 638 | 2473 | 833 |
| UNS | 0.39 | 0.36 | 0.38 | 717 | 1119 | 1989 | 1267 |

**Table 2**: *Average performance of the basic loanword detection model as compared to other models, with bold indicating the best-performing model for each metric*

Although the model's superiority is reflected in its higher F1 score, the absolute magnitude of this score can still be characterized as modest. Nevertheless, performance patterns suggest a non-monotonic relationship between the proportion of loanwords in the dataset and overall model effectiveness. This trend is particularly evident in the per-language evaluation, where the model's performance varies in accordance with the density of loanwords in the respective monolingual corpora.

|  | PREC. | RECALL | F1 | LOAN N° |
| --- | --- | --- | --- | --- |
| ENGLISH | 0.85 | 0.79 | 0.82 | 414 |
| GERMAN | 0.66 | 0.69 | 0.67 | 382 |
| PORTUGUESE | 0.61 | 0.63 | 0.62 | 332 |
| SPANISH | 0.57 | 0.69 | 0.63 | 311 |
| FRENCH | 0.60 | 0.66 | 0.63 | 292 |
| ITALIAN | 0.49 | 0.75 | 0.60 | 253 |

**Table 3**: *Relationship between the number of loanwords in data and F1 score obtained during loanword detection in each monolingual wordlist*

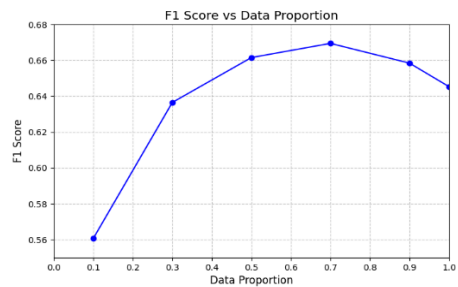

**Figure 3.** *F1 score as a function of training data proportion.*

Apart from the Portuguese dataset, there is a general trend that languages with more loanwords tend to yield higher F1 scores. As shown in Figure 3, when varying the proportion of data (while preserving the relative frequency of loanwords), model performance initially improves with more data, reaching a peak before slightly declining. This suggests that increasing data quantity can enhance performance, likely because it provides the model with richer and more representative linguistic patterns. However, beyond a certain threshold, the benefits diminish, possibly due to noise effects. More importantly, the plateau and slight decline in F1 at higher data proportions suggest that simply adding more data is not always beneficial; the quality, diversity, and representativeness of that data play an equally important role. This implies that targeted data selection and not indiscriminate data expansion is desired.

Performance improvements were also observed when the model was augmented with cross-linguistic information. Table 4 presents the model's performance under this condition. Consistent with previous findings, the Germanic languages (English and German), which contain the highest proportion of loanwords in their datasets, achieved the best performance, outperforming the Romance languages. However, in contrast to the results obtained with the basic model, the Romance language data in this configuration also exhibited a positive correlation between the number of loanwords and model performance.

|  | PRECISION | RECALL | F1 | TP | FP | TN | FN |
|---|---|---|---|---|---|---|---|
| ENGLISH | 0.81 | 0.91 | 0.86 | 378 | 86 | 337 | 36 |
| GERMAN | 0.67 | 0.91 | 0.77 | 347 | 174 | 328 | 35 |
| PORTUGUESE | 0.59 | 0.90 | 0.71 | 299 | 208 | 304 | 33 |
| SPANISH | 0.55 | 0.92 | 0.69 | 285 | 235 | 297 | 26 |
| FRENCH | 0.57 | 0.85 | 0.68 | 247 | 185 | 364 | 45 |
| ITALIAN | 0.46 | 0.96 | 0.62 | 243 | 286 | 304 | 10 |
| AGGREGATED | 0.61 | 0.91 | 0.73 | 1799 | 1174 | 1934 | 185 |

**Table 4.** *Performance of the model augmented with cross-linguistic information, compared to the basic. Results are reported per language, with loanword frequency shown to highlight its correlation with model accuracy.*

Notably, the cross-linguistically scaled model achieved, on average, a 6% performance increase relative to the basic model. This finding suggests that further scaling, particularly by incorporating semantic information, where available, has the potential to yield substantial additional improvements in performance.

### 7.1. Error analysis

We focus our error analysis on the basic loanword detection model, as it constitutes the architectural core of the scaled variant. Understanding error patterns at this foundational level offers insights into the root causes of misclassifications, which may propagate in the scaled configurations.

An analysis of false positive classifications revealed several systematic error patterns in the basic model. First, the model frequently misclassified high-frequency core vocabulary items, e.g., *arm* [German, Spanish], *back* [German, Spanish, Italian], *brother* [English, German, Italian, Portuguese], and *child* [English, French, German, Spanish], as loanwords. The model also exhibited a tendency to over-identify widely shared internationalisms and scientific or technical terms (e.g., *energy* [French, Spanish, Italian, Portuguese], *university* [French, Spanish, Italian, Portuguese], *president* [French, Spanish, Italian, Portuguese], *computer* [English, French]) as loanwords across all target languages. A recurring characteristic among these misclassifications is the presence of consonant clusters, a feature that, while not explicitly modelled, appears to be implicitly learned and overgeneralized. Additional examples include *blues* [English], *passacaglia* [French, Italian], *grunge* [English], *silhouette* [French], *capriccio* [French], *allegro* [Italian], *wanderlust* [German], *peligro* [Spanish], and *sangre* [Spanish].

Beyond consonant clusters, the model appears sensitive to other phonotactic and phonetic cues. In French, words with nasal vowels – when transcribed (e.g., *restaurant* [rɛstorã] 'restaurant', *liaison* [lieɪzɔ̃] 'liaison', *enseignant* [ãsɛɲã] 'teacher', *genre* [ʒɑ̃ʁ] 'genre') and rounded vowels (e.g., *cheveux* [ʃəvø] 'hair', *mur* [myʁ] 'wall', *œil* [œj] 'eye') were disproportionately misclassified as loanwords. In Spanish, similar errors occurred with words containing velar fricatives (e.g., *pájaro* [paxaro] 'bird', *bolígrafo* [boliɣrafo] 'pen', *fuego* [fweɣo] 'fire'). Although such segmental and phonotactic features were not explicitly included in the feature representation, the model appears to capture them implicitly, leading to overgeneralization. This likely explains why the performance of an ablated model, augmented with explicit segmental and phonotactic features, was lower than that of the original model, as the same cues may have been redundantly represented, thereby introducing noise.

False negatives also exhibited systematic patterns. Many highly integrated borrowings, particularly from Romance languages, were misclassified as native words. These items often had simple consonant-vowel syllable structures (e.g., *opera*, *sofa*, *pasta*). This phonotactic simplicity appears to override other cues in the model's predictions, even for borrowings from non-Romance languages, such as *sushi*, *kimono*, *kabuki*, and *fado*. Furthermore, false negatives were, on average, shorter in length than false positives, indicating that the model may have internalized the statistical tendency for native words to be shorter than borrowings, despite the inclusion of dynamic length features.

Overall, these patterns indicate that the model captures a range of phonotactic and phonetic cues beyond those explicitly encoded. While this implicit feature learning is advantageous, it does result in systematic misclassifications. Controlling for such overgeneralized features, either through feature selection or targeted normalization, represents a potential avenue for improving model performance.

## 8. Conclusion

In this study, we introduced a feature-rich, statistically-grounded autonomous unsupervised model for loanword detection, designed for integration into historical linguistics workflows. Unlike most prior approaches, which rely primarily on language-external information, our model is driven by features extracted exclusively from the internal linguistic features of the language data under investigation. Specifically, the model incorporates phonological, phonotactic and morphological features, with statistical controls for word length, syntactic category, and morphological patterns. These features are iteratively modeled until convergence is achieved.

Evaluation on a dataset comprising Germanic languages (English, German) and Romance languages (French, Italian, Spanish, Portuguese) shows that our approach substantially outperforms the existing baseline. This suggests that, when linguistically relevant features are embedded within an appropriate statistical architecture, the model can detect loanwords even when they are deeply integrated into the recipient language. Furthermore, the system exhibits self-learning properties, enabling it to generalize beyond explicitly encoded patterns.

Moreover, the model's architecture is inherently scalable. Incorporating cross-linguistic information yields further performance gains, and future extensions may incorporate more precise control of phonotactic constraints, modeling for language-specific phonetic properties, and the integration of semantic embeddings to identify semantic fields with high loanword probability.

While our evaluation was limited to six languages, the model's design is extensible to larger datasets and bigger language families. Its capacity to unify feature extraction and statistical modeling makes it especially promising for low-resource contexts, where language-contact histories and genealogical relationships are unknown. For such cases, our framework offers a flexible, interpretable, and empirically validated baseline for automated loanword detection.

**Acknowledgements**
This research is being funded by a grant from the European Research Council (ERC) under the European Union's Horizon Europe Framework Program (horizon) grant number 101045195.

Special thanks to all members of the BANG project for their contributions, especially Aurore Montébran, who sadly passed away before the model was completed, for her constant encouragement.

**References**

Alvarez-Mellado, E., Porta-Zamorano, J., Lignos, C., & Gonzalo, J. (2025). *Overview of ADoBo at IberLEF 2025 : Automatic Detection of Anglicisms in Spanish* (arXiv:2507.21813).

Crawford, C. J. (2009). *Adaptation and transmission in Japanese loanword phonology*. Cornell University.

Grant, A. P. (2000). Fabric, pattern, shift and diffusion: What change in Oregon Penutian languages can tell historical linguists. *Survey Reports, Survey of California and Other Indian Languages*.

Grant, A. P. (2015). Lexical Borrowing. In J. R. Taylor (Éd.), *The Oxford Handbook of the Word*. Oxford University Press.

Hafez, O. (1996). Phonological and morphological integration of loanwords into Egyptian Arabic. *Égypte/Monde Arabe*, *27-28*, 383-410.

Hantgan, A., & List, J.-M. (2022). Bangime : Secret language, language isolate, or language island? A computer-assisted case study. *Papers in Historical Phonology*, *7*, 1-43.

Haspelmath, M. (2009). Lexical borrowing: Concepts and issues. *Loanwords in the world's language: A Comparative Handbook*, 35-54.

Heath, J. (2022). Origins of Dogon NP tonosyntax. *Diachronica*, *39*(5), 707-741.

Kang, Y. (2011). Loanword phonology. *The Blackwell companion to phonology*, *4*, 1003-1026.

Lehmann, W. P. (1962). *Historical linguistics: An introduction*. Rinehart & Winson.

List, J.-M. (2012a). LexStat: Automatic detection of cognates in multilingual wordlists. *Proceedings of the EACL 2012 Joint Workshop of LINGVIS & UNCLH*, 117-125.

List, J.-M. (2012b). SCA: Phonetic Alignment Based on Sound Classes. In D. Lassiter & M. Slavkovik (Éds.), *New Directions in Logic, Language and Computation* (Vol. 7415, p. 32-51). Springer Berlin Heidelberg.

List, J.-M. (2019a). Automatic detection of borrowing (Open problems in computational diversity linguistics 2). In *The Genealogical World of Phylogenetic Networks*. Blogger.

List, J.-M. (2019b). Statistical proof of language relatedness (Open problems in computational diversity linguistics 7). *The Genealogical World of Phylogenetic Networks.*, *6*(8), 1-6.

List, J.-M. (2024). Open problems in computational historical linguistics. *Open Research Europe*, *3*, 201.

List, J.-M., & Forkel, R. (2022). Automated identification of borrowings in multilingual wordlists. *Open Research Europe*, *1*, 79.

Mennecier, P., Nerbonne, J., Heyer, E., & Manni, F. (2016). A Central Asian language survey: Collecting data, measuring relatedness and detecting loans. *Language Dynamics and Change*, *6*(1), 57-98.

Miller, J. E., Tresoldi, T., Zariquiey, R., Beltran Castanon, C. A., Morozova, N., & List, J.-M. (2020). Using lexical language models to detect borrowings in monolingual wordlists. *Plos one*, *15*(12), e0242709.

Moran, S., & Prokić, J. (2013). Investigating the relatedness of the endangered Dogon languages. *Literary and linguistic computing*, *28*(4), 676-691.

Paradis, C., & LaCharité, D. (2008). Apparent phonetic approximation: English loanwords in Old Quebec French1. *Journal of Linguistics*, *44*(1), 87-128.

Poplack, S., & Dion, N. (2012). Myths and facts about loanword development. *Language variation and change*, *24*(3), 279-315.

Prakhya, S., & P, D. (2020). *Unsupervised Separation of Native and Loanwords for Malayalam and Telugu* (arXiv:2002.05527). arXiv.

Ralli, A., & Rouvalis, A. (2022). Morphological Integration of Loan Words in Kaliardá. *Languages*, *7*(3), 167.

Rama, T., & List, J.-M. (2019). An automated framework for fast cognate detection and Bayesian phylogenetic inference in computational historical linguistics. Proceedings of the 57th Annual Meeting of the Association for Computational Linguistics, 6225–6235.

Swadesh, M. (1955). Towards Greater Accuracy in Lexicostatistic Dating. *International Journal of American Linguistics*, *21*(2), 121-137.

Tadmor, U., Haspelmath, M., & Taylor, B. (2010). Borrowability and the notion of basic vocabulary. *Diachronica*, *27*(2), 226-246.

Thomason, S. G., & Kaufman, T. (2001). *Language contact* (Vol. 22). Edinburgh University

# Appendix

## A1. Some Asymmetric Classifications

Both the basic and scaled models exhibit stronger performance on datasets from Germanic languages than on those from Romance languages. An important observation is that whenever the basic model classifies a token as a loanword, the scaled model also assigns it to the loanword category. The converse, however, does not hold: instances identified as loanwords by the scaled model may be classified as native words by the basic model. This asymmetry indicates that the scaled model constitutes a refinement of the basic model's predictions. Overall, the scaled model consistently outperforms the basic model in classification accuracy. Table 1 presents representative cases in which the scaled model correctly labels a token as a loanword, whereas the basic model misclassifies it as native.

| WORD | LANGUAGE | BASIC_PREDICTION | SCALED_PREDICTION | TRUE_LABEL |
|---|---|---|---|---|
| zero | english | native | loanword | loanword |
| opera | english | native | loanword | loanword |
| karma | english | native | loanword | loanword |
| shampoo | german | native | loanword | loanword |
| guru | german | native | loanword | loanword |
| mojo | german | native | loanword | loanword |
| ninja | french | native | loanword | loanword |
| futon | french | native | loanword | loanword |
| ramen | french | native | loanword | loanword |
| azúcar | spanish | native | loanword | loanword |
| sofá | spanish | native | loanword | loanword |
| espresso | spanish | native | loanword | loanword |
| foto | italian | native | loanword | loanword |
| gin | italian | native | loanword | loanword |
| brandy | italian | native | loanword | loanword |
| salsa | portuguese | native | loanword | loanword |
| rumba | portuguese | native | loanword | loanword |
| mambo | portuguese | native | loanword | loanword |

**Table 1.** Examples of asymmetric classifications where the scaled model correctly identifies loanwords that the basic model misclassifies as native, illustrating the scaled model's higher recall and refinement over the basic model.

## A2. Hyperparameters

The model consists of three principal components: a feature extractor, a scoring engine, and an iterative detection mechanism, with each governed by a set of hyperparameters established through preliminary, iterative experimentation. During feature extraction, the model computes character n-grams for all lengths between two and ten characters, a range selected to capture both short and extended patterns relevant to the detection of loanwords. N-grams occurring with a frequency below 0.005 are assigned a strong penalty, while those below 0.02 receive a smaller penalty, thereby emphasizing sequences atypical of the native lexicon. A similar approach is applied to bigram probabilities, with values below 0.01 attracting high penalties and those below 0.05 lower penalties, in order to capture uncommon phoneme-to-phoneme transitions often present in borrowings. The feature set also includes two entropy measures i.e., n-gram entropy and transition entropy, that represent the distributional unpredictability of sequences; both measures are normalized prior to scoring. Word length on the other hand, is standardized as a dynamically calculated z-score relative to the input set, with large deviations from the mean length serving as a potential indicator of borrowed origin.

The scoring engine integrates these normalized features through a weighted combination, with the rare n-gram score contributing the most, followed in influence by the rare transition score, transition entropy, n-gram entropy, average transition probability, and length z-score. Part-of-speech weights are then applied to account for the varying likelihoods of borrowing among word classes: nouns receive no adjustment, adjectives are reduced by half, verbs by 70%, adverbs by 80%, and function words by 95%. These adjusted scores are transformed via a sigmoid function into borrowing probabilities, with boosts applied when certain rarity thresholds are exceeded.

In the iterative detection phase, words with a borrowing probability of at least 0.3 are provisionally classified as borrowed. The model then re-runs on the remaining unclassified items for a maximum of seven iterations, progressively refining results as high-confidence borrowings are removed. Iteration halts early if the change in the borrowed set between consecutive steps falls below 1% of its size, indicating convergence. From the second iteration onward, the model applies additional pattern-based refinements, adjusting probabilities based on differences in prefix, suffix, and internal phonemic frequencies between words classified as native and borrowed.

### A3. Extended results

To enable a more comprehensive evaluation of our model's performance, in addition to standard performance metrics presented in Table xx, we present the f1 scores per language in Table 2 below. Again, compared to the ablated versions, and the baseline model, our model outperforms on all language datasets.

|  | ENGLISH | GERMAN | PORTUGUESE | SPANISH | FRENCH | ITALIAN |
| --- | --- | --- | --- | --- | --- | --- |
| AUT_BOR | 0.81 | 0.67 | 0.62 | 0.63 | 0.62 | 0.60 |
| AUT_BOR-P1 | 0.36 | 0.26 | 0.19 | 0.29 | 0.15 | 0.30 |
| AUT_BOR-P2 | 0.78 | 0.53 | 0.54 | 0.52 | 0.54 | 0.46 |
| AUT_BOR-AUG | 0.79 | 0.62 | 0.61 | 0.55 | 0.56 | 0.47 |
| UNS | 0.20 | 0.53 | 0.42 | 0.43 | 0.22 | 0.35 |

**Table 2**: Comparison of per-language F1 scores across the proposed basic model, ablated configurations, and the baseline model. Higher scores indicate better performance.